\pdfoutput=1

\documentclass[11pt]{article}

\usepackage[]{ACL2023}
\usepackage{multirow}
\usepackage{booktabs}
\usepackage{graphicx}
\usepackage[table,xcdraw]{xcolor}
\usepackage{tabularx}
\usepackage{times}
\usepackage{latexsym}
\usepackage{float}        
\usepackage{longtable}    
\usepackage{caption}      
\usepackage{tcolorbox}
\tcbuselibrary{breakable}
\newtcolorbox{promptbox}[1]{colback=gray!5, colframe=black!60,
    title={#1}, fonttitle=\bfseries\small, left=2mm, right=2mm}
\usepackage{float}  
\usepackage{amsmath}
\usepackage{amssymb}
\usepackage{tcolorbox}
\usepackage{bbm}
\usepackage{threeparttable}
\usepackage[T1]{fontenc}

\usepackage[utf8]{inputenc}

\usepackage{microtype}

\usepackage{inconsolata}

\usepackage{subcaption}
\usepackage{enumitem}

\newcommand{\nonspotrandom}{Weakly-annotated}
\newcommand{\spot}{Feature-annotated}
\newcommand{\spotlc}{feature-annotated}
\newcommand{\nonspotrandomlc}{weakly-annotated}

\title{Evidence-Anchored Overdose Death Prediction from Death Investigation Reports}

\author{Raavi Gupta$^{1}$, Ziqing Yang$^{1}$, \textbf{Karli R. Hochstatter}$^2$, Talia Nadel$^2$, \\
\textbf{Nicole D'Anna}$^3$, \textbf{Jason K. Graham}$^3$, \textbf{Hannah Johnson}$^4$, \textbf{Kelsey Opozda}$^3$, \\
\textbf{Nabila El-Bassel}$^{1}$, \textbf{Anubhav Jangra}$^{1}$, \textbf{Smaranda Muresan}$^{5}$\\
$^1$Columbia University, USA, $^2$Friends Research Institute, USA\\
$^3$The New York City Office of Chief Medical Examiner, USA \\  
$^{4}$Independent Researcher, USA, $^5$Barnard College, Columbia University, USA
}

\begin{document}
\maketitle
\begingroup
\def\thefootnote{}\footnotetext{Corresponding authors: \texttt{raavi02g@gmail.com}, \texttt{anubhav@cs.columbia.edu} and \texttt{smuresan@barnard.edu}.}
\endgroup
\begin{abstract}
Confirmation of overdose deaths in the United States is often delayed due to the complexities of certifying drug-involved deaths. Death investigation reports represent a rich but underutilized data source that could be used to predict whether a death is a suspected overdose (i.e., a death preliminarily identified as an overdose prior to the final cause and manner of death determination). We present an automated framework that classifies suspected overdoses from death investigation reports, while producing supporting explanations guided by expert knowledge. Through an expert-guided design process, we investigate 1) what type of explanations are best for model prediction performance, and 2) which architecture is better suited for this task: pipeline models that rationalize first and then predict, or self-rationalizing models that jointly predict and rationalize? Our experiments show that a pipeline model with explanations grounded in the input death investigation reports outperforms the other settings. Our best model also outperforms a rule-based system currently in deployment by 15\%, while allowing expert interventions for final decisions. 

\end{abstract}

\section{Introduction}

The age-adjusted rate of overdose death in the US jumped from 8.9 per 100,000 population in 2003 to 31.3 in 2023 \cite{Kochanek2024databriefs}. Lag time for drug overdose deaths is generally longer than that for all other causes of death by more than six months \cite{cdc2025vsrr}, and the resulting delays prevent rapid surveillance and obstruct timely public health interventions \cite{gelberg2024need}. An underutilized source of early information lies in the narrative reports produced by on-scene death investigators. These reports are typically available shortly after a fatality and contain rich information about physical findings, substance involvement, and witness observations \cite{davis2020position, odonnell2022using}.

\begin{figure}[t]
    \centering
\includegraphics[width=\linewidth]{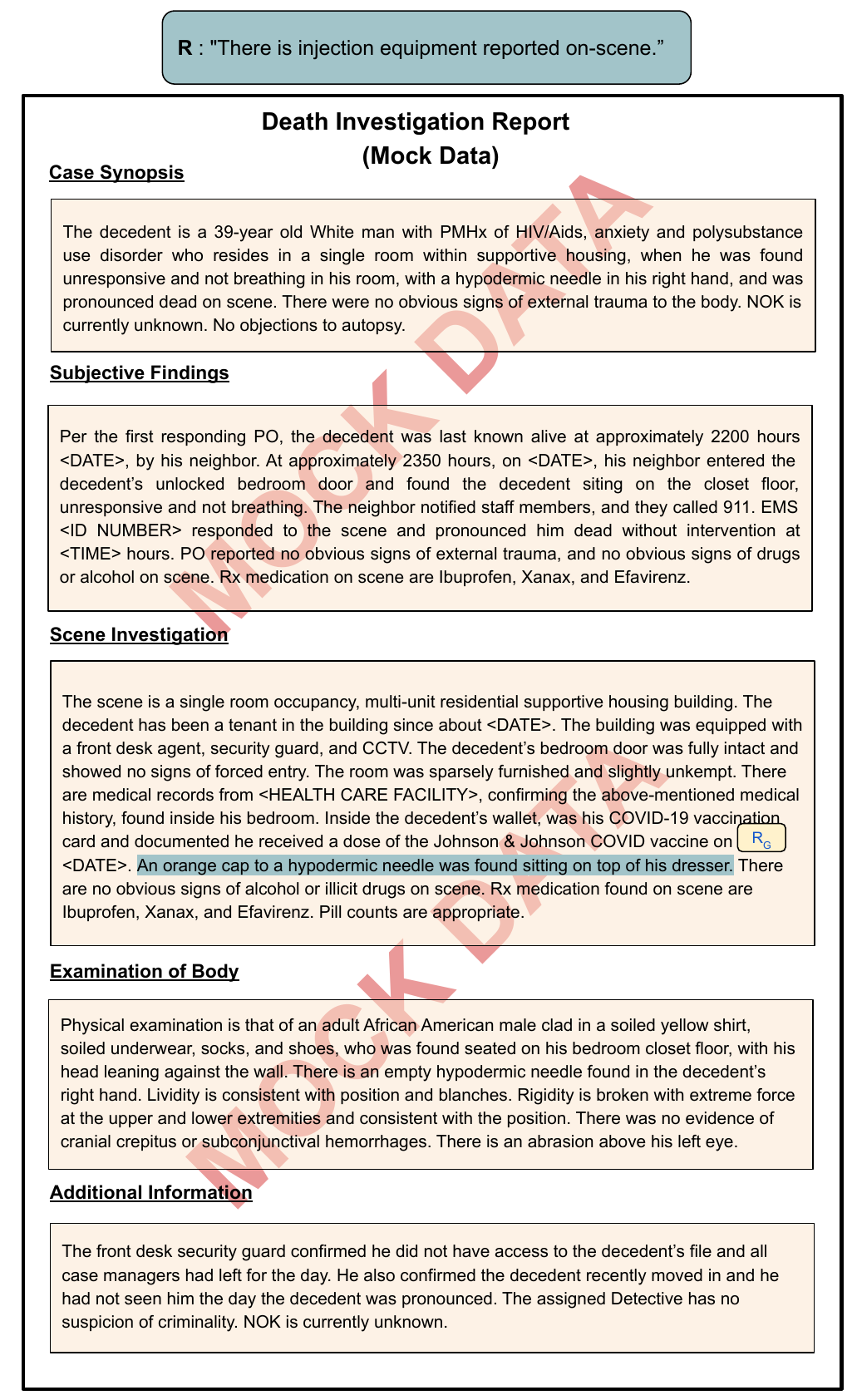}
    \caption{The two explanation types on an example death investigation report of mock data (completely deidentified). Feature-verbalizations $R$ (top) are predefined statements per expert-identified feature; feature-guided rationales $R_G$ (highlighted) are verbatim sentences extracted from the report corresponding to the given expert-identified feature.}
    \label{fig:rationales}
\end{figure}

Existing early-detection systems depend on manual investigator input, ranging from a single suspected-overdose flag \cite{hochstatter2024rapid} to pre-defined features extracted from the reports \cite{Hochstatter2022SPOT, vannice2022accuracy}, making these systems time-intensive and dependent on a small, highly trained workforce. Feature-based ML pipelines \cite{goodmanmeza2022development} and traditional NLP \cite{ward2019enhancing} have been used to automate parts of this process, but these methods operate over death certificates, which may not be available until six months after the death \cite{spencer2017timeliness}. 

\begin{figure*}[t]
    \includegraphics[width=\linewidth]{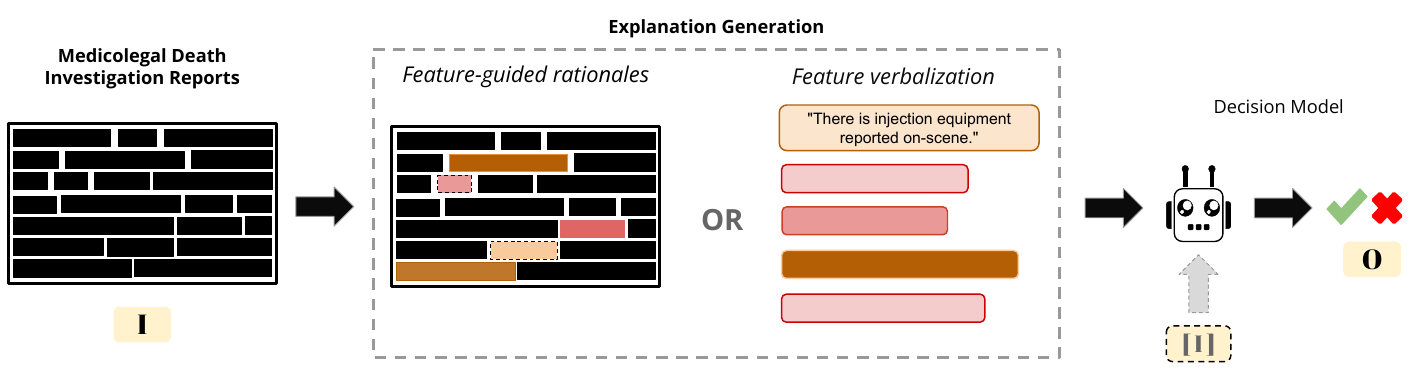}
    \caption{Overview of the proposed method. We compare two types of explanations: feature-guided rationales $R_G$ (sentences extracted from the report) and feature-verbalizations $R$ (predefined statements for each present feature). A classifier model then predicts the overdose label $O$ from the explanation, together with the report. We show here the decoupled regime, in which explanation generation and prediction are two separate steps rather than a single joint generation.
    } \label{fig:intro_fig}
\end{figure*}




While Large language models (LLM) have revolutionized NLP applications in various domains, 
their risks of hallucinations \cite{turpin2023languagemodelsdontsay, Huang_2025}, non-deterministic output \cite{he2025nondeterminism}, and black box nature requires care in applications in high-stake domains such as overdose prediction. 
To overcome these limitations, we adopt an expert-guided design process \cite{cho2026expertguidedclinicaltextaugmentation}, directly collaborating with domain experts to explore the use of open-source LLMs for predicting overdose death directly from death investigation reports, while producing supporting explanations guided by expert knowledge. 

We investigate two questions: 1) what type of explanations are best for model prediction performance and 2) are pipeline models that rationalize first and then predict or self-rationalizing models that jointly predict and rationalize \cite{wiegreffe-etal-2021-measuring} better for this task.
To answer the first question, 
we explore two types of 
explanations: 1) \textit{feature verbalization}, which are natural language descriptions of the expert-provided features associated with a death investigation report, and 2) \textit{feature-guided rationales}, which are explicit text spans extracted from the death investigation reports that correspond to an expert-provided feature. Figure \ref{fig:rationales} shows an expert-written mock death investigation report and the corresponding rationales obtained by the overdose prediction system: $R$ represents the feature verbalization for the feature "Injection Equipment on Scene", while $R_{G}$ shows the exact text span from the death investigation report corresponding to this expert-provided feature (i.e., feature-guided rationale). 





Our results show that a pipeline model with feature-guided rationales using open-source models\footnote{We focus on open-source models, as proprietary models may raise data privacy and HIPAA compliance concerns in clinical settings with sensitive information.} is the best setup for this tasks. Moreover, our best model achieves a 15\% improvement in overdose prediction over an adopted expert rule-based system that use death investigation reports for overdose predictions \cite{Hochstatter2022SPOT}. The pipeline model also allows experts to refine explanations during inference rather than starting from scratch, which could lead to substantial time savings per report. We make three key contributions in this work:
\begin{itemize}[noitemsep]
    \item To the best of our knowledge, we propose the first end-to-end system for automated decision support in overdose prediction from death investigation reports, eliminating the need for manual data extraction and enabling the system to scale.
    \item We present an interpretable approach, supporting an expert-guided design in which feedback drives model refinement during both training and inference.
    \item  We conduct a systematic comparison of explanation types (feature verbalization and feature-guided rationales) and model types (pipeline and self-rationalizing models) for overdose prediction. 
\end{itemize}
\section{Related Work}

\subsection{Explanation generation}

Early work on model interpretability focused on post-hoc explanation methods such as attention visualization \citep{bahdanau2016neuralmachinetranslationjointly}, saliency maps, and LIME-style perturbation approaches  \citep{ribeiro2016why}. However, these methods do not provide human-readable rationales that domain experts can provide feedback upon. \citet{lei2016rationalizingneuralpredictions} pioneered the extraction of human-readable rationales from the input that are sufficient to support the model's predictions.

The advent of LLMs significantly improved rationale generation quality. \citet{wei2022chain} demonstrated that chain of thought prompting that produces intermediate reasoning steps before answering improves performance on tasks. \citet{sun2022investigatingbenefitsfreeformrationales}  found that incorporating rationales to even a small amount of training data can considerably improve model performance. Subsequent work, however, examined whether such rationales reflect the model's true reasoning or merely serve as post-hoc justifications \citep{turpin2023languagemodelsdontsay}, motivating approaches that grounds rationales in input evidence rather than generating them freely.  \citet{wiegreffe-etal-2021-measuring} formalized this concern and proposed association tests to assess whether a model's rationale and prediction are genuinely linked. They further unveiled that free-form rationales are not faithful by construction. Extractive rationale, on the other hand, constrains explanations to verbatim spans of the source document \cite{wei-jie-etal-2024-plausible} and makes the reasoning auditable, unlike free-form rationales. 

Expert feedback has emerged as a mechanism for improving model-generated rationales. \citet{saakyan-muresan-2024-iclef} introduced an in-context learning framework that integrates scarce human annotations into model training to generate high-quality explainable outputs, illustrating how expert feedback loops can enhance the accuracy of the model. \citet{hase-etal-2024-unreasonable} demonstrated that LLMs can generalize from easy annotated data to harder instances via lightweight fine-tuning, suggesting that carefully curated supervision signals can substitute for large-scale human annotation.

\subsection{Medical NLP and Explanation-Driven Decision Support}

Medical NLP has a history of extracting structured signals from free text, spanning rule-based systems \citep{friedman2004automated} and, more recently, transformer-based models such as BioBERT \citep{lee2020biobert} and ClinicalBERT \citep{huang2020clinicalbertmodelingclinicalnotes}, fine-tuned specifically on biomedical corpora.  \citet{mcinerney-etal-2024-towards} proposed an interpretable risk prediction model that uses LLMs to surface evidence passages in electronic health records supporting diagnostic hypotheses, demonstrating that evidence-aligned predictions reduce diagnostic delays. 

Early approaches to overdose classification utilize traditional ML \citep{goodmanmeza2022development, tang2023natural} techniques on death certificates. More recent work has demonstrated the use of LLMs for structured information extraction from clinical free text with high precision \citep{wiest2024privacy, funnell2026improving}.

Moreover, prior work has focused primarily on analysis from death certificates, which may not be available for more than six months. Our work operate on death investigation reports, which are available within days of death rather than months enabling rapid identification of overdose deaths. \citet{Hochstatter2022SPOT} and \citet{harruff2023medical} also use death investigation reports to identify suspected overdose deaths. However, their rule-based systems require extensive expert annotation. Other systems such as  \citet{hochstatter2024rapid} output only an expert provided global flag based on the totality of evidence, discarding the criterion-level evidence that our automated and intervenable framework preserves.

\section{Methods and Experimental Setting}

Let $\textbf{I}$ denote the input death investigation report, $\textbf{O} \in \{0, 1\}$ the suspected overdose prediction, and $V = \{v_1, \ldots, v_{|V|}\}$ the set of expert-defined features. We also define the explanations $\mathbf{R}$ that justifies the prediction $\textbf{O}$. The explanations are human-readable and each explanation map back to a feature $v_i$.

We define two forms of explanations - \textit{feature-verbalizations} ($R$): natural language template mapped to the expert-provided features $v_i$ and \textit{feature-guided rationales} ($R_G$): extracted evidence spans corresponding to the expert-defined features $v_i$. An example of the two types of explanation is provided in Figure \ref{fig:rationales}. Table \ref{tab:feature-verbalized-statements} in Appendix E shows all the expert-defined features predictive of overdose death that can be associated with a death investigation report (e.g., History of Prior Overdose(s), Benzodiazepine(s), Pills/Capsules on Scene). We use $\mathbf{R} \in \{R, R_G\}$ to refer to either form when a statement applies to both.


\subsection{Dataset} \label{sec: dataset}

Our dataset's structure informs the design of our proposed methodology, including the explanation generation and prediction models. The dataset comprises of 9,263 death investigation reports with two splits:
\begin{itemize}[noitemsep]
    \item \textbf{\spot}: Reports for which expert-labeled boolean features $\{v_k\}$ and the final overdose labels $\mathbf{O}$ are available. 
    \item \textbf{\nonspotrandom}: Reports for which only the final overdose labels $\mathbf{O}$ are available, without feature annotations $\{v_k\}$. 
\end{itemize}

\begin{table}[t]
\centering
\caption{Overdose label distribution across the \spot\ and \nonspotrandom\ partitions. \textit{Yes} and \textit{No} denote whether the death was certified as an overdose.}
\label{tab:data-distribution}
\resizebox{\linewidth}{!}{%
\begin{tabular}{llrrr}
\toprule
\textbf{Partition} & \textbf{Set} & \textbf{No} & \textbf{Yes} & \textbf{Total} \\
\midrule
\multirow{3}{*}{\spot}
& Train & 786   & 3{,}697 & 4{,}483 \\
& Val   & 95    & 457     & 552 \\
& Test  & 96    & 449     & 545 \\
\midrule
\multirow{3}{*}{\nonspotrandom}
& Train & 1{,}828 & 1{,}099 & 2{,}927 \\
& Val   & 229     & 145     & 374 \\
& Test  & 233     & 149     & 382 \\
\bottomrule
\end{tabular}
}
\end{table}

Table~\ref{tab:data-distribution} presents the distribution of overdose labels across the train, validation, and test sets, created by random sampling from the full pool of reports. Overall, 64.7\% of reports are labeled as overdoses; given this class imbalance, we report macro-F1 throughout. Moreover, 39.8\% of reports lack expert feature annotations, motivating our use of models to infer them and further obtain downstream explanations $\mathbf{R}$.

\subsection{Explanation Generation and Prediction Models} \label{sec:explanation-generation}
Our dataset contains the binary values for each feature only for the feature-annotated partition. To augment the weakly-annotated partition with these missing annotations, we used five-shot question answering with \texttt{Llama-3.1-8B} \cite{grattafiori2024llama3herdmodels} to extract feature labels. The system prompt, per-feature questions, and configuration details are provided in Appendix~\ref{app:feature-extraction}. Table \ref{tab:feature-extraction} provides the per-feature extraction performance for the feature-guided rationales.
We used this fully augmented feature-labeled dataset to generate the two types of explanations. We denote the completed feature value by $\hat{v}_k \in \{0, 1\}$, which corresponds to the annotated value $v_k$ when available and otherwise to the feature value predicted by Llama-3.1-8B.

\subsubsection{Feature Verbalization} \label{subsec:feature-verbalized}
Each variable $v_k$ (e.g., ``Drugs/Paraphernalia: Injection Equipment on Scene'') is verbalized to a statement $s_k$ (e.g., ``There is injection equipment reported on-scene.''); the full set of mappings is provided in Table~\ref{tab:feature-verbalized-statements}. The feature-verbalized explanation is then constructed by enumerating the statements for all variables indicated as present:
\[
R = \{ s_k \mid \hat{v}_k = 1, k = 1, \ldots, n \}
\]

We obtained the feature indicators from two sources: for the \spotlc\ partition, we used the expert-labeled indicators directly, whereas for the \nonspotrandomlc\ partition, we inferred them using \texttt{Llama-3.1-8B}, as described in Section \ref{sec:explanation-generation}. The indicators were then mapped to their corresponding statements $s_k$ and concatenated to form the feature-verbalized explanations.

\paragraph{Prediction Models.}
We considered two paradigms for producing the final prediction from the constructed explanations. 
    \textbf{Pipeline} ($I \rightarrow R$; $I\,R \rightarrow O$), where the explanation generation and prediction are separated into two stages. The first model generates the explanation from the input report, and the second model then takes the explanation, alongside the original report ($I$), as input to produce the final prediction.
    \textbf{Self-rationalizing} ($I \rightarrow R, O$): A single model receives the input report and produces both the explanation and the final prediction in a single pass.
We trained two decoder-only LLMs for both pipeline and self-rationalize models ($\mathbf{I} \rightarrow \mathbf{O}$, $\mathbf{I} \rightarrow R,\mathbf{O}$, and $R \rightarrow \mathbf{O}$), namely \texttt{Llama-3.1-8B Instruct}~\cite{grattafiori2024llama3herdmodels} and \texttt{Qwen3-8B}~\cite{qwen3technicalreport}. These models are of comparable size but differ in their pretraining emphasis: \texttt{Llama-3.1} provides a strong general-purpose baseline, while \texttt{Qwen3} offers strong performance on instruction-following tasks. We fine-tuned each model with LoRA \cite{hu2022lora} on the constructed explanations and gold output labels.

\subsubsection{Feature-guided Rationales} \label{subsec:feature-guided}
Feature-guided rationales are extracted directly from the death investigation report where each rationale is a verbatim sentence-level span\footnote{Our design intentionally extracts rationales at the sentence level rather than as arbitrary spans for simplicity and robustness. Unbounded spans introduce additional noise, along with limiting the contextual knowledge from shorter spans, making the experts read the full sentence regardless of span width.} that can be traced back to the specific part of the report. 
We cast this rationale extraction as a multi-label sentence classification task. Let $\mathbf{I} = \{x_1, \ldots, x_m\}$ denote the $m$ sentences of a report. For each sentence-variable pair $(x_i, v_k)$, we predict a binary label $\hat{r}^G_{i,k} \in \{0, 1\}$ indicating whether $x_i$ constitutes evidence for feature $v_k$. The feature-guided rationale is the set of sentences supporting at least one variable:
\[
R_G = \{\, x_i \mid \exists\, v_k \in V : \hat{r}^G_{i,k} = 1 \,\}
\]

Our dataset contains no gold-labeled evidence spans for either partition. We therefore construct silver-standard labels for $\hat{r}^G_{i,k}$ over the full dataset via a teacher--student pipeline. We use an ensemble of teacher LLMs' labels for each sentence--feature pair independently, and aggregate their outputs by majority vote to obtain the silver labeled $R_G^{silver}$ for each report. The teacher ensemble used to construct silver labels comprises MedGemma-27B~\cite{sellergren2026medgemmatechnicalreport}, 
Llama-70B~\cite{grattafiori2024llama3herdmodels}, and Qwen3-30B~\cite{qwen3technicalreport}, inferred in full precision.

These silver labels are then used to fine-tune a student encoder-only model,
 for sentence-level multi-label classification
predicting $\hat{r}^G_{i,k}$ for all features $v_k \in V$ simultaneously given $\mathbf{I}$. At inference, the fine-tuned student extracts $R_G$ directly from $\mathbf{I}$, without the teacher ensemble. For the student model, we used \texttt{Clinical-ModernBERT} ~\cite{lee2025clinicalmodernbertefficientlong}, a lightweight encoder-only model pre-trained on medical corpora. 

\paragraph{Prediction Models.}
Since these explanations are extracted by an encoder-only model, self-rationalization 
is not applicable for this explanation type, and we therefore consider only the pipeline models.
\textbf{Pipeline ($I \rightarrow R_G;\ [I]R_G \rightarrow O$):} the paradigm in which a student model first extracts grounded rationales $R_G$ from $I$, and a separate classifier then predicts $O$ from $R_G$ alone or from $(I, R_G)$ jointly, ablating the contribution of the original report at the classification stage. 
We additionally investigate whether rationales alone are sufficient for classification, \textit{viz.,} whether ${R_G}  \rightarrow O$ can match the performance of a model with full access to the original report. 
This motivates a direct comparison between $R_G\ \rightarrow O$ and $I \rightarrow O$, isolating the information contributed by the extracted 
rationales versus the original narrative. 

\subsubsection{Implementation Details}

All decoder-only models are fine-tuned using LoRA
~\cite{hu2022lora} with rank $r = 32$ and scaling factor $\alpha = 32$, applied to all attention and feed-forward projection layers on single NVIDIA A100-40GB GPU. Models are loaded in 4-bit quantization and fine-tuned with gradient checkpointing enabled. The maximum sequence length is set to 4,096 tokens to accommodate longer reasoning traces. Inference is performed using vLLM~\cite{kwon2023efficientmemorymanagementlarge} for efficient generation.
\texttt{Clinical-ModernBERT} is fine-tuned in full precision for the sentence-level multi-label classification task, with a classification head.

\section{Results}

Table~\ref{tab:main_results} reports accuracy and macro-F1 across all model variants and evaluation splits. The rule-based baseline with expert-human annotations, SPOT \cite{Hochstatter2022SPOT}, achieves 0.721 Macro-F1. However, in contrast, the best overall configuration of our proposed approach (Qwen3 with $I, R_G \rightarrow O$), achieves 0.831 Macro-F1 without any expert knowledge. This configuration illustrates that grounded rationales together with the original report result in the strongest prediction signal. We examine the results through three views: the explanation type, pipeline versus self-rationalizing models, and the data splits.

\begin{table*}[t]
  \begin{center}
  \small
  \caption{Comparison of all variants across test splits.
    $I \rightarrow O$: direct prediction from report.
    $\ R_G \rightarrow O$: pipeline model with just feature-guided rationale as input for prediction.
    $I, R_G \rightarrow O$: pipeline model with feature-guided rationales and report as input.
    $I \rightarrow R, O$: self-rationalized model with feature verbalization as explanations.
    $I, R \rightarrow O$: pipeline model with report and feature verbalization as explanations  as input for prediction.
    \textbf{Bold} = best configuration per row.}
  \label{tab:main_results}
  \resizebox{\textwidth}{!}{%
  \begin{tabular}{@{} ll cc cc cc cc cc @{}}
    \toprule
    & & \multicolumn{2}{c}{$I \rightarrow O$}
      & \multicolumn{2}{c}{$\ R_G \rightarrow O$}
      & \multicolumn{2}{c}{$I, R_G \rightarrow O$}
      & \multicolumn{2}{c}{$I \rightarrow R, O$}
      & \multicolumn{2}{c}{$I, R \rightarrow O$} \\
    \cmidrule(lr){3-4} \cmidrule(lr){5-6} \cmidrule(lr){7-8}
    \cmidrule(lr){9-10} \cmidrule(lr){11-12}
    \textbf{Split} & \textbf{Model}
      & \textbf{Acc.} & \textbf{Macro-F1}
      & \textbf{Acc.} & \textbf{Macro-F1}
      & \textbf{Acc.} & \textbf{Macro-F1}
      & \textbf{Acc.} & \textbf{Macro-F1}
      & \textbf{Acc.} & \textbf{Macro-F1} \\
    \midrule
    \multirow{2}{*}{\textbf{Overall}}
      & \texttt{Llama-3.1} & 0.832 & 0.811 & 0.798 & 0.778 & \textbf{0.847} & \textbf{0.829} & 0.809 & 0.779 & 0.832 & 0.812 \\
      & \texttt{Qwen3}  & 0.789 & 0.766 & 0.786 & 0.765 & \textbf{0.849} & \textbf{0.831} & 0.799 & 0.768 & 0.794 & 0.776 \\
    \midrule
    \multirow{2}{*}{\textbf{\spot}}
      & \texttt{Llama-3.1} & 0.851 & 0.674 & 0.828 & 0.656 & \textbf{0.866} & \textbf{0.728} & 0.837 & 0.619 & 0.842 & 0.639 \\
      & \texttt{Qwen3}  & 0.820 & 0.637 & 0.818 & 0.632 & \textbf{0.873} & \textbf{0.738} & 0.829 & 0.597 & 0.828 & 0.605 \\
    \midrule
    \multirow{2}{*}{\textbf{\nonspotrandom}}
      & \texttt{Llama-3.1} & 0.804 & 0.791 & 0.757 & 0.736 &\textbf{0.819}  & \textbf{0.809}& 0.770 & 0.762  & 0.817 & 0.801 \\
      & \texttt{Qwen3}  & 0.743 & 0.725 & 0.741 &0.717  &\textbf{0.814}  & \textbf{0.803} &0.757 &0.747 & 0.746 & 0.705   \\
      \bottomrule
  \end{tabular}}
  \end{center}
  \vspace{0.5em}
  \small{We compute SPOT's \citep{Hochstatter2022SPOT} performance on our test split obtaining 0.736 accuracy and 0.721 macro-F1.}
\end{table*}

\paragraph{Analysis 1: Explanation types} 
Feature-guided rationales outperform feature-verbalized explanations on every split for both models, with the best feature-guided model exceeding the best feature-verbalized model by 2.3\% macro-F1. Apart from final prediction accuracy,  we also compare per-feature
extraction quality on the \spotlc{} split for the feature-guided rationale type and find that across all the features the model extracted features match the human extracted features around 75\% of the cases and there is a discrepancy of at least one feature in the remaining 25\%. (More details in Appendix \ref{app:feature-extraction}). We also test whether just indicating the presence of feature alone carries the signal required to predict overdose and subsequently trained a logistic regression classifier on the binary variable indicators. The best logistic regression model trained using feature-guided rationales achieved 0.621 macro-F1 which is substantially lower than the corresponding best Qwen3 model by 33.8\%. Therefore, knowing which features are present is not sufficient; the rationale text and report are needed to predict overdose outcomes, underscoring the value of feature-guided rationales over feature-verbalized explanations and bare feature presence.

\paragraph{Analysis 2: Pipeline vs Self-rationalizing Models.}
For feature-verbalized explanations, the pipeline regime outperforms the self-rationalizing regime for both Llama-3.1 (0.812 vs.\ 0.779 macro-F1) and Qwen3 (0.776 vs.\ 0.768), i.e., conditioning the prediction on the report together with a pre-generated explanation is more effective than generating the explanation and prediction in a single pass. This comparison is only possible for feature-verbalized explanations, since feature-guided rationales use an encoder-only student that cannot generate explanation and prediction jointly. \citet{wiegreffe-etal-2021-measuring} report the opposite trend, where self-rationalizing models outperform pipeline models. Their pipeline, however, predicts from the explanation alone, and they show that explanations are often not sufficient without the original input. Our pipeline gives the classifier both the death investigation report and the explanation, which avoids this limitation. Our results for feature-guided rationales also confirm their finding on sufficiency: the rationale alone is not sufficient, as $R_G \rightarrow O$ performs significantly worse than both $I \rightarrow O$ and $I, R_G \rightarrow O$.

\paragraph{Analysis 3: Across different data splits.}
The macro-F1 for \nonspotrandomlc\ split is higher than the corresponding \spotlc\ split for the best model which is surprising given the training data in the \spotlc\ split is human labeled. One of the possible explanations for this can be that almost all samples in the \spotlc\ split contain features that are indicative of overdose and hence distinguishing between them is harder than \nonspotrandomlc\ split despite having human-labeled data. Another possible reason can be due to the class imbalance in the \spot\ dataset.

\begin{table*}[t]
\caption{F1 scores for sentence-level classification without and with expert feedback integration. \% Improvement is computed as $\frac{\text{w/ Feedback} - \text{w/o Feedback}}{\text{w/o Feedback}} \times 100$. "S.U.H." denotes "Substance Use History" and "D/P" denotes "Drugs/Paraphernalia".}
\label{tab:f1_scores}
\centering
\resizebox{0.82\linewidth}{!}{
\begin{tabular}{l|ccc}
\toprule
\textbf{Variable} & \textbf{w/ Feedback} & \textbf{w/o Feedback} & \textbf{\% Improv.} \\
\midrule
History of Prior Overdose(s)                                    & \textbf{0.5882} & 0.5195          & +13.2  \\
S.U.H.: Benzodiazepine(s)                        & \textbf{0.5185} & 0.2632          & +97.0  \\
S.U.H.: Cocaine (including Crack)                & 0.8673          & \textbf{0.8689} & $-$0.2 \\
S.U.H.: Injection Drug Use                       & \textbf{0.5116} & 0.4835          & +5.8   \\
S.U.H.: Ketamine                                 & \textbf{1.0000} & \textbf{1.0000} & 0.0    \\
S.U.H.: Opioid(s)                                & \textbf{0.7820} & 0.7070          & +10.6  \\
S.U.H.: PCP                                      & 0.0000          & 0.0000          & \textemdash \\
S.U.H.: Prescription Medication(s)               & \textbf{0.3243} & 0.0000          & \textemdash \\
S.U.H.: Methamphetamine/Other Stimulant(s)       & \textbf{0.6222} & 0.3030          & +105.3 \\
S.U.H.: Unspecified/Other                        & \textbf{0.7985} & 0.7845          & +1.8   \\
Substance Use Treatment History                                 & \textbf{0.7864} & 0.7311          & +7.6   \\
D/P: Injection Equipment on Scene               & 0.7869          & \textbf{0.8176} & $-$3.8 \\
D/P: Other Paraphernalia on Scene               & \textbf{0.6559} & 0.4797          & +36.7  \\
D/P: Packaging on Scene                         & \textbf{0.8145} & 0.7821          & +4.1   \\
D/P: Pills/Capsules on Scene                    & \textbf{0.8247} & 0.6990          & +18.0  \\
D/P: Powder/Residue on Scene                    & \textbf{0.8070} & 0.7539          & +7.0   \\
D/P: Smoking Equipment on Scene                 & 0.7729          & \textbf{0.7914} & $-$2.3 \\
D/P: Straw or Similar on Scene                  & 0.8451          & \textbf{0.8488} & $-$0.4 \\
\midrule
\textbf{Aggregate}                                              & \textbf{0.6836} & 0.6018          & +13.6  \\
\bottomrule
\end{tabular}
}
\end{table*}

\subsection{Teacher Agreement for Silver Data Construction}

We analyzed the pairwise Cohen's $\kappa$ agreement between the three teacher models used to construct the silver-labeled dataset for feature-guided rationale extraction $R_G$ (see Table \ref{tab:kappa-scores}). Aggregate three-way agreement between all the teacher models is moderate ($\kappa = 0.361$), with MedGemma-LLaMa showing the strongest pairwise agreement ($\kappa = 0.500$) and Qwen3-LLaMa the weakest ($\kappa = 0.284$).

\begin{table*}[t]
\centering
\small
\setlength{\tabcolsep}{5pt}
\caption{Qualitative analysis of three representative cases under the decoupled paradigm. Evidence spans are extracted verbatim and abridged here. Cases (a) and (b) show correct extraction of evidence; case (c) shows failure propagating from the extraction stage.} \label{tab:case-study}
\renewcommand{\arraystretch}{1.15}
\begin{tabular}{@{}p{0.075\textwidth} p{0.30\textwidth} p{0.28\textwidth} p{0.25\textwidth}@{}}
\toprule
 & \textbf{(a) Correct rejection}
 & \textbf{(b) Missed suspected overdose}
 & \textbf{(c) Extraction error} \\
 & Gold: No \quad Pred: No~\checkmark
 & Gold: Yes \quad Pred: No~$\times$
 & Gold: Yes \quad Pred: No~$\times$\\
\midrule

\rotatebox[origin=c]{0}{\textbf{Evidence}} &
\raggedright
$\bullet$ 
person with PMHx of CAD, HTN, DM, COPD, \emph{remote h/o drug use}, depression and schizophrenia; found down in a shelter.\newline
$\bullet$ The only drug use she is aware of is marijuana use \emph{in their teens--20s}.
&
\raggedright
$\bullet$ Drank socially; \emph{family denied} any known illicit drug use.\newline
$\bullet$ A \emph{cut red straw} and a blue pill (imprint B706) found on a nearby table.\newline
$\bullet$ \emph{Powder, appearing to be crushed pills}, in the clear plastic wrapper of a cigarette box.
&
\raggedright
$\bullet$ No illicit drugs were found.\newline
$\bullet$ Family stated the decedent was not known to use drugs or alcohol.
\tabularnewline

\midrule
\rotatebox[origin=c]{0}{\textbf{$\hat{v}_k$}} &
\raggedright
History of substance use: \textbf{1}\newline
Drug paraphernalia: 0
&
\raggedright
History of substance use: 0\newline
Paraphernalia at scene: \textbf{1}\newline
Loose/non-prescribed pills: \textbf{1}
&
\raggedright
History of substance use: 0\newline
Paraphernalia at scene: 0\newline
Loose/non-prescribed pills: 0
\tabularnewline

\midrule
\rotatebox[origin=c]{0}{\textbf{Analysis}} &
\raggedright
The model correctly classifies this case as a non-overdose. 
A history-of-substance-use flag is present, but the extracted spans reveal that the “remote h/o drug use” refers specifically to marijuana. Since marijuana is excluded from the substance use history in the overdose prediction problem due to low risk, this is a false annotation that is correctly rejected using the additional context from the extracted evidence.
&
\raggedright
The model fails to classify this case as a suspected overdose despite
extracting substantial scene evidence: drug packaging (a clear plastic
wrapper), a loose pill not prescribed to the decedent, a cut straw, and
powder/residue. The rationale is accurate, but the classifier does not
weigh the accumulated scene indicators strongly enough against the family's
denial of drug use.
&
\raggedright
Here the error originates in the extraction stage: the model extracts
misleading spans, which propagate to an incorrect classification. This
failure mode is specific to the decoupled paradigm and marks the point
where expert review of the rationale prior to prediction would intervene and help in downstream classification.
\tabularnewline
\bottomrule
\end{tabular}
\end{table*}

We found that the agreement varies considerably across features, which also correlates with annotation difficulty. Scene evidence variables (\textit{e.g.,} \textit{Powder/Residue on Scene} ($\kappa = 0.497$) and \textit{Packaging on Scene}) ($\kappa = 0.433$) that are easy to annotate tend to have higher agreement as compared to substance use history features, particularly \textit{Prescription Medication(s)} ($\kappa = 0.171$). This pattern is consistent with the F1 results in Table~\ref{tab:f1_scores}, where the same feature showed the largest gains from expert feedback (see Section \ref{sec: expert-driven}), suggesting that low inter-teacher agreement is a reliable signal of features requiring targeted 
correction.

\subsection{Expert driven online testing} \label{sec: expert-driven} 
We fine-tuned \texttt{ClinicalModernBERT} on the silver-labeled dataset for multi-label sentence classification across the features in $V$ as described in \ref{subsec:feature-guided}, producing the student model for feature-guided rationale extraction $R_G$. Table~\ref{tab:f1_scores} reports per-feature F1 scores for extraction performance on the \spotlc\ split with and without expert feedback integration. The aggregate F1 improves from 0.602 without feedback to 0.684 with feedback, a 13.6\% gain.

Expert feedback yields the most substantial improvements for rare or underspecified features (eg. \textit{Prescription Medication(s)} and \textit{Benzodiazepines}). These improvements highlight the value of domain expertise in disambiguating forensic terminology. Conversely, well-represented features such as \textit{Cocaine} and \textit{Ketamine} remain stable, indicating that the teacher ensemble already provides strong silver supervision for these categories. \textit{PCP} achieves 0.000 F1 in both conditions, likely due to extreme rarity in the dataset. For physically observable scene evidence such as \textit{Injection Equipment} and \textit{Straw}, feedback marginally hurts performance which may be attributed to noise.

We provide the taxonomy of types of major errors found from expert feedback in Appendix \ref{sec:expert-feedback}.

\section{Case Study}
To characterize where the pipeline paradigm succeeds and where it does not, we examine three representative cases from the \spotlc\ test set (Table~\ref{tab:case-study}). Together they separate two different failure modes: errors of aggregation, where evidence is correctly
surfaced but incorrectly weighed, and errors of extraction, where the
evidence never reaches the classifier at all and hence leads to error in the downstream prediction task as well.

\section{Discussion}

In our experiments, feature-guided rationales consistently outperform feature verbalizations across all data splits, which highlights the value of grounding explanations in exact spans of the input. Although feature-guided rationales alone ($R_G \rightarrow O$) do not match the full report ($I \rightarrow O$; Table~\ref{tab:main_results}), they complement it: when we condition on both $I$ and $R_G$, we obtain our best configuration (0.831 macro-F1). This signal lies in the evidence text rather than in the features it represents, as a logistic regression over the same features reaches only 0.621 macro-F1. In case (a) of Table~\ref{tab:case-study}, for instance, the rationale reveals that a flagged substance use history refers only to marijuana, which allows the model to correctly reject the case.

Our pipeline decouples explanation generation from prediction, which allows experts to correct the explanation behind each feature before it reaches the classifier. When we incorporate expert feedback, rationale extraction F1 improves by 13.6\% (Table~\ref{tab:f1_scores}). In our study, experts most often remove on-scene evidence that the model reads as a history of use and add synonyms that the model misses. Without these corrections, such errors propagate to the final prediction (Table~\ref{tab:case-study}, case c).

Prior work on overdose prediction either treats the model as a black-box labeler~\citep{goodmanmeza2022development, funnell2026improving} or depends on manual annotation of death investigation reports~\citep{Hochstatter2022SPOT, harruff2023medical}. Our framework requires no manual annotation at inference, allows experts to refine the evidence behind each prediction, and uses only open-weight models that run locally, so sensitive data never leaves the practitioner's environment.





\section{Conclusion}
The practical bottleneck our system targets is timeliness. Finalized overdose mortality data are delayed by months because confirmation depends on toxicology findings, autopsy reports, and sometimes other investigative information. Whereas, the deployed alternatives that work on death investigation reports for identifying suspected overdose deaths depend on meticulous manual annotation that scales poorly.
Our paper presented a study that investigated what types of explanations and what type of explanation-guided models (pipeline or self-rationalizing) are best for this task, showing that a pipeline approach with feature-guided rationales grounded in the death investigation report is the best setup for this task. 
This setup enables an expert 
to refine machine-extracted rationales rather than reading each report from
scratch,  offering an earlier signal at lower human cost and without asking public-health practitioners to trust an opaque model. The correctable explanation can increase the trustworthiness of the system: it keeps a human accountable for each determination while removing the manual-extraction burden that currently limits surveillance throughput. We see this as the central contribution for the public-health community, not just a faster model but a model whose outputs domain experts can review and adjust as needed to ensure validity within their framework.

\section*{Limitations}
Our evaluation is conducted on a single jurisdiction, and it remains an open question how well the learned models transfer to coroner or medical examiner offices with different documentation practices. The expert-defined features, while operationally validated, may not capture the full range of overdose-relevant evidence across jurisdictions or evolving drug landscapes. Additionally, the silver-labeled training data for feature-guided rationale extraction is inherently noisy,  moderate aggregate teacher agreement ($\kappa = 0.361$) means that a fraction of sentence-level labels are incorrect, potentially introducing systematic errors for the lowest-agreement variables. Finally, the expert feedback loop we evaluate is retrospective; prospective deployment would require careful interface design to ensure that corrections are both efficient and consistent across reviewers.
\section*{Ethical considerations}
Robust protocols were put in place to prevent the release of confidential information. This study was approved by our institutions IRB. 
All data sharing 
occurred via a secure SharePoint environment that had been configured to be both Health Insurance Portability and Accountability Act (HIPAA) and Food and Drug Administration (FDA) compliant. Model training and development occurred within a certified cloud-based environment 
approved for use with Personally Identifiable Information (PII), Personal Health Information (PHI), and HIPAA data. Before the research team had access to the data, narrative death investigation reports were de-identified using the Presidio analyzer to prevent the unnecessary use and release of identifiable data during AI model training and development. 

\section*{AI use disclosure}
We have used generative AI tools to aid in writing the code for metric generation and generating tables. We have not used AI tools in our core experiment implementation and have also refrained from using them for ideation, literature review, figure generation, or end-to-end paper or section writing. We have reviewed all AI-assisted work, validated the code used for metric generation, cross-checked the illustration artifacts, and ensured that all authors had full agency over the paper-writing process. We take responsibility for the final content of this work, including text, claims, or artifacts produced with the aid of generative AI.

\bibliography{refs}
\bibliographystyle{acl_natbib}

\newpage
\appendix

\section{Feature Extraction for the \nonspotrandom\ Partition}
\label{app:feature-extraction}

For each feature, we paired the death investigation report with a question asking whether
that variable was present, and the model answered Yes or No. The prompt
template is shown in Figure~\ref{fig:prompt-feature-extraction}. The average macro-F1 across the features for the \spotlc\ split is 0.967. More detailed results can be found in Table \ref{tab:feature-extraction}.

\begin{table*}[t]
\centering
\caption{Feature extraction accuracy for on \spot\ test split. H = human; AI = our system} \label{tab:feature-extraction}
\resizebox{0.75\textwidth}{!}{%
\begin{threeparttable}
\begin{tabular}{lcccccccc}
\toprule
\textbf{Variable} & \textbf{Pos\textsubscript{H}} & \textbf{Pos\textsubscript{AI}} & \textbf{TP} & \textbf{FP} & \textbf{FN} & \textbf{TN} & \textbf{Recall} & $\boldsymbol{\kappa}$ \\
\midrule
\multicolumn{9}{l}{\textit{Substance use history}} \\
History of prior overdose(s)        & 45  & 41  & 41  & 0 & 4   & 500 & 0.911 & 0.950 \\
Benzodiazepine(s)                   & 25  & 22  & 22  & 0 & 3   & 520 & 0.880 & 0.933 \\
Cocaine (including Crack)           & 173 & 173 & 173 & 0 & 0   & 372 & 1.000 & 1.000 \\
Injection drug use                  & 48  & 47  & 47  & 0 & 1   & 497 & 0.979 & 0.988 \\
Ketamine                            & 2   & 2   & 2   & 0 & 0   & 543 & 1.000 & 1.000 \\
Opioid(s)                           & 164 & 160 & 160 & 0 & 4   & 381 & 0.976 & 0.982 \\
PCP                                 & 3   & 2   & 2   & 0 & 1   & 542 & 0.667 & 0.799 \\
Other Rx used not as prescribed     & 10  & 7   & 7   & 0 & 3   & 535 & 0.700 & 0.821 \\
Methamphetamine / other stimulant   & 14  & 14  & 14  & 0 & 0   & 531 & 1.000 & 1.000 \\
Unspecified / other                 & 367 & 338 & 338 & 0 & 29  & 178 & 0.921 & 0.884 \\
Treatment history                   & 96  & 92  & 92  & 0 & 4   & 449 & 0.958 & 0.974 \\
\midrule
\multicolumn{9}{l}{\textit{Drugs / paraphernalia on scene}} \\
Injection equipment                 & 53  & 52  & 52  & 0 & 1   & 492 & 0.981 & 0.989 \\
Other paraphernalia\tnote{*}        & 205 & 103 & 103 & 0 & 102 & 340 & 0.502 & 0.558 \\
Packaging                           & 120 & 120 & 120 & 0 & 0   & 425 & 1.000 & 1.000 \\
Pills / capsules                    & 44  & 40  & 40  & 0 & 4   & 501 & 0.909 & 0.948 \\
Powder / residue                    & 133 & 133 & 133 & 0 & 0   & 412 & 1.000 & 1.000 \\
Smoking equipment                   & 89  & 89  & 89  & 0 & 0   & 456 & 1.000 & 1.000 \\
Straw or similar                    & 71  & 70  & 70  & 0 & 1   & 474 & 0.986 & 0.992 \\
\bottomrule
\end{tabular}
\end{threeparttable}
}
\end{table*}

\begin{figure}[t]
\centering
\begin{promptbox}{Prompt: Five-shot feature extraction}
\small\ttfamily
\textbf{System Prompt.}\\[2pt]
You are a Medical Investigator that answers questions based on provided context.
Answer the question `Yes' or `No' with direct evidence from the context.\\[4pt]
Do not infer specific drugs from generic mentions of terms like `substance use disorder', `illicit substance' etc.
For questions about history of substance use, answer `Yes' only if the text explicitly mentions past use of that substance regardless of equipment found on scene.
For questions about equipment found on scene, answer `Yes' only if the text explicitly mentions the presence of that equipment on the scene and not based on inferences.
Do not infer information.\\[6pt]
\textbf{User Input.}\\[2pt]
Examples:\\
\{Feature specific examples\}\\
Context:\\
\{report\}\\[4pt]
Question:\\
\{feature\_question\}\\[6pt]
\textbf{Output.}\\[2pt]
Yes / No
\end{promptbox}
\caption{Prompt used for five-shot feature indicator extraction with
Llama-3.1-8B-Instruct.}
\label{fig:prompt-feature-extraction}
\end{figure}

\subsection{Generation Setting}
\label{app:generation-setting}

We extract features with Llama-3.1-8B-Instruct using greedy decoding with five in-context examples per feature. Each question is posed separately against each section of the report, and the feature indicator $\hat{v}_k$ is set to present if any section yields an affirmative response.

\subsection{Question for Each Feature}
\label{app:feature-questions}

Table~\ref{tab:feature-questions} lists the question $q_k$ paired with each
feature variable $v_k$ in the five-shot extraction prompt described above.

\begin{table*}[p]
\centering
\small
\caption{Questions used to extract each feature variable from the MLI
report. Each question is substituted into the
\texttt{\{feature\_question\}} slot of the prompt template.}
\label{tab:feature-questions}
\resizebox{0.8\linewidth}{!}{
\begin{tabular}{>{\raggedright\arraybackslash}p{0.24\textwidth}>{\raggedright\arraybackslash}p{0.69\textwidth}}
\toprule
\textbf{Variable} & \textbf{Question ($q_k$)} \\
\midrule
History of Prior Overdose(s) & Is there any mention of a prior overdose? \\
Benzodiazepine(s) & Is there any reported substance use history of benzodiazepines? \\
Cocaine (including crack) & Is there any reported substance use history of cocaine (including crack)? \\
Injection Drug Use & Is there any reported substance use history of injection drug use (answer No if only the word paraphernalia is present without any further evidence)? \\
Ketamine & Is there any reported substance use history of ketamine? \\
Opioid(s) & Is there any reported substance use history of opioid(s)? \\
PCP & Is there any reported substance use history of phencyclidine (PCP)? \\
Prescription Medication(s) & Is there any reported substance use history of prescription medications, excluding specified opioids, benzodiazepines, or stimulants? \\
Methamphetamine\slash Other Stimulant(s) & Is there any reported substance use history of methamphetamine? \\
Unspecified/Other & Is there any reported substance use history of substance(s) other than benzodiazepines, cocaine, injections, ketamine, opioids, PCP, methamphetamine (answer No if only the words `illicit substance' etc.\ are present without naming the substance)? \\
Substance Use Treatment History & Is there any record of past substance use treatment by a professional? \\
Injection Equipment on Scene & Is there any indicator of injection equipment reported on-scene, including syringes, needle caps, cookers, spoons, tourniquets (answer No if only the word paraphernalia is present without any further evidence)? \\
Other Paraphernalia on Scene & Is there any indicator of evidence of paraphernalia (excluding syringes, needle caps, cookers, spoons, tourniquets) reported on-scene, such as scales, fentanyl test strips (answer No if only the word paraphernalia is present without any further evidence)? \\
Packaging on Scene & Was any drug packaging, such as glassines, baggies, or flip-top vials, reported on the scene, regardless of whether drugs were present? \\
Pills/Capsules on Scene & Is there any indicator of loose pills/capsules reported on-scene, not prescribed to the decedent? \\
Powder/Residue on Scene & Is there any indicator of powder or residue reported on-scene (solid capsules/pills are not considered as powder/residue)? \\
Smoking Equipment on Scene & Is there any indicator of smoking equipment reported on-scene, including glass pipes, filters, thin rods, lighters (not alone)? \\
Straw or Similar on Scene & Is there any indicator of straw or similar equipment reported on-scene, including cut straws, rolled paper (including currency), deconstructed pens? \\
\bottomrule
\end{tabular}
}
\end{table*}

\section{Feature-verbalized Statements}
\label{app:feature-verbalized-statements}

Table~\ref{tab:feature-verbalized-statements} lists the statements used to
construct feature-verbalized explanations $R$. For each feature $v_k$
predicted as present ($\hat{v}_k = 1$), the corresponding statement $s_k$
is included in the explanation.
\begin{table*}[ht]
\centering
\small
\caption{Predefined natural-language statements mapped to each of the feature variables, used to construct the feature-verbalized explanations $R$.}
\label{tab:feature-verbalized-statements}
\resizebox{0.8\linewidth}{!}{
\begin{tabular}{>{\raggedright\arraybackslash}p{0.25\textwidth}>{\raggedright\arraybackslash}p{0.68\textwidth}}
\toprule
\textbf{Feature} & \textbf{Statement ($s_k$)} \\
\midrule
History of Prior Overdose(s) & There is mention of a prior overdose. \\
Benzodiazepine(s) & There is reported substance use history of benzodiazepines. \\
Cocaine (including Crack) & There is reported substance use history of cocaine. \\
Injection Drug Use & There is reported substance use history of IV injection drugs. \\
Ketamine & There is reported substance use history of ketamine. \\
Opioid(s) & There is reported substance use history of opioid(s). \\
PCP & There is reported substance use history of PCP. \\
Prescription Medication(s) & There is a reported history of prescription medicines, other than opioids, benzodiazepines, or stimulants, not being used as prescribed. \\
Methamphetamine\slash Other Stimulant(s) & There is reported substance use history of methamphetamine. \\
Unspecified/Other & There is reported substance use history of unspecified substance(s). \\
Substance Use Treatment History & There is substance use treatment history by a professional. \\
Injection Equipment on Scene & There is injection equipment reported on-scene. \\
Other Paraphernalia on Scene & There is paraphernalia (excluding injection equipment) reported on-scene. \\
Packaging on Scene & There is drug packaging such as glassines, baggies, or flip-top vials reported on-scene. \\
Pills/Capsules on Scene & There are loose pills/capsules/tablets not prescribed to the decedent reported on-scene. \\
Powder/Residue on Scene & There is powder/residue reported on-scene. \\
Smoking Equipment on Scene & There is smoking equipment such as glass pipes or filters reported on-scene. \\
Straw or Similar on Scene & There are straws or similar equipment like rolled paper or deconstructed pens reported on-scene. \\
\bottomrule
\end{tabular}
}
\end{table*}



\section{Teacher Ensemble Prompt}
\label{app:prompt-teacher}

Figure~\ref{fig:prompt-rationale-extraction} shows the prompt used by each
teacher model to extract feature-guided rationales via question answering.
Prompts are reproduced verbatim from our experiments.

\begin{figure*}[t]
\centering
\resizebox{0.8\linewidth}{!}{
\begin{promptbox}{Prompt: Feature-guided rationale extraction}
\small\ttfamily
Given the following investigative report, answer the question by providing verbatim evidence from the text.\\
\{report\}\\
Question: \{question\}\\
\{instruction\}\\[4pt]
You are a medicolegal death investigator that answers questions strictly based on the provided report. Follow these rules:\\[2pt]
\textbf{Evidence Only:} Output only sentences extracted verbatim from the report.\\[2pt]
\textbf{No Inference:} Do not infer specific substances from generic mentions (e.g., ``substance use disorder,'' ``illicit substances'') or from equipment/paraphernalia unless explicitly stated.\\[2pt]
\textbf{Substance Use:} Answer ``history of substance use'' questions only if the report explicitly mentions past misuse of the specific substance asked in the question. Presence of the substance (or something suspected to be the substance) on the scene is not sufficient to indicate a history of misuse of the substance. For prescription medications, there must some mention of use not as directed or prescribed; a history of taking or having been prescribed a medication with no context or evidence to suggest it was misused is not sufficient.\\[2pt]
\textbf{Equipment:} Answer ``equipment and paraphernalia found on scene'' questions only if the report explicitly mentions the presence of the specified equipment or paraphernalia.\\[2pt]
\textbf{Unanswerable:} If no explicit evidence is found, output only: Unanswerable.\\[2pt]
\textbf{Multiple Statements:} If multiple disjointed pieces of text from the report provide evidence, enumerate only the minimal relevant fragments (not full sentences). Each fragment must be copied verbatim from the report and clearly enumerated.\\[2pt]
Final responses must consist of only the supported pieces of text from the investigative report (or Unanswerable) answering the question and \textbf{no extra information} that does not answer the question.
\end{promptbox}
}
\caption{Prompt used for extracting feature-guided rationales via question
answering in teacher models.}
\label{fig:prompt-rationale-extraction}
\end{figure*}

\subsection{Teacher Agreement}
\label{app:teacher-agreement}

Pairwise agreement among the three teacher models is reported in Table \ref{tab:kappa-scores}.
\begin{table*}[]
\caption{Cohen's $\kappa$ agreement between teacher models for sentence-level evidence extraction across 18 forensic variables. Values represent pairwise and three-way agreement in the silver dataset construction.} \label{tab:kappa-scores}
\centering
\resizebox{\textwidth}{!}{%
\begin{tabular}{lllll}
\toprule
\textbf{Variable}                                                  & 
\textbf{MedGemma and Qwen3} & \textbf{MedGemma and Llama} & \textbf{Qwen3 and Llama} & \textbf{All Annotators} \\
\toprule
\textbf{History of Prior Overdose(s)}                              & 0.406112                   & 0.656521                    & 0.391848                & 0.484827                \\
\textbf{Substance Use History: Benzodiazepine(s)}                  & 0.251875                   & 0.423022                    & 0.265501                & 0.313466                \\
\textbf{Substance Use History: Cocaine (including Crack)}          & 0.326150                   & 0.562726                    & 0.347551                & 0.412142                \\
\textbf{Substance Use History: Injection Drug Use}                 & 0.131958                   & 0.404878                    & 0.104612                & 0.213816                \\
\textbf{Substance Use History: Ketamine}                           & 0.297646                   & 0.620090                    & 0.324882                & 0.414206                \\
\textbf{Substance Use History: Opioid(s)}                          & 0.285187                   & 0.504151                    & 0.319806                & 0.369715                \\
\textbf{Substance Use History: PCP}                                & 0.335513                   & 0.508541                    & 0.315829                & 0.386628                \\
\textbf{Substance Use History: Prescription Medication(s)}         & 0.120104                   & 0.286659                    & 0.105637                & 0.170800                \\
\textbf{Substance Use History: Methamphetamine/Other Stimulant(s)} & 0.172003                   & 0.397143                    & 0.144348                & 0.237832                \\
\textbf{Substance Use History: Unspecified/Other}                  & 0.234560                   & 0.505633                    & 0.214964                & 0.318385                \\
\textbf{Substance Use Treatment History}                           & 0.347706                   & 0.555369                    & 0.338634                & 0.413903                \\
\textbf{Drugs/Paraphernalia: Injection Equipment on Scene}         & 0.397759                   & 0.513754                    & 0.311943                & 0.407819                \\
\textbf{Drugs/Paraphernalia: Other Paraphernalia on Scene}         & 0.205337                   & 0.327110                    & 0.187710                & 0.240052                \\
\textbf{Drugs/Paraphernalia: Packaging on Scene}                   & 0.383756                   & 0.569978                    & 0.347442                & 0.433726                \\
\textbf{Drugs/Paraphernalia: Pills/Capsules on Scene}              & 0.230831                   & 0.492686                    & 0.225899                & 0.316472                \\
\textbf{Drugs/Paraphernalia: Powder/Residue on Scene}              & 0.424763                   & 0.607269                    & 0.460214                & 0.497416                \\
\textbf{Drugs/Paraphernalia: Smoking Equipment on Scene}           & 0.423799                   & 0.521117                    & 0.326593                & 0.423836                \\
\textbf{Drugs/Paraphernalia: Straw or Similar on Scene}            & 0.396304                   & 0.543575                    & 0.378239                & 0.439373                \\
\midrule
\textbf{Aggregate}                                                 & 0.298409                   & 0.500012                    & 0.283981                & 0.360801  \\
\bottomrule
\end{tabular}}
\end{table*}

\subsection{Teacher Accuracy}
\label{app:teacher-accuracy}

Teacher accuracy against expert feature labels is reported in Table \ref{tab:teacher_extraction_accuracy}.
\begin{table*}[ht]
\centering
\small
\caption{Per-variable extraction accuracy of three teacher models against human-annotated labels
across all \spot\ cases ($n = 5{,}580$; train\,/\,val\,/\,test\,=\,4{,}483\,/\,552\,/\,545).
A variable is predicted positive if the model's extracted span list is non-empty.
PPV\,=\,1.000 for all variables and all models (zero false positives) and is omitted for brevity.
Cohen's $\kappa$ is computed against human labels as ground truth.}
\label{tab:teacher_extraction_accuracy}
\resizebox{0.9\textwidth}{!}{%
\begin{tabular}{@{}lr ccc ccc ccc@{}}
\toprule
& & \multicolumn{3}{c}{\textbf{Llama-70B}} 
  & \multicolumn{3}{c}{\textbf{MedGemma-27B}} 
  & \multicolumn{3}{c}{\textbf{Qwen3-30B}} \\
\cmidrule(lr){3-5} \cmidrule(lr){6-8} \cmidrule(lr){9-11}
\textbf{Variable} & \textbf{Pos\textsubscript{H}}
  & \textbf{Pos\textsubscript{AI}} & \textbf{Recall} & \textbf{$\kappa$}
  & \textbf{Pos\textsubscript{AI}} & \textbf{Recall} & \textbf{$\kappa$}
  & \textbf{Pos\textsubscript{AI}} & \textbf{Recall} & \textbf{$\kappa$} \\
\midrule
\multicolumn{11}{@{}l}{\textit{Substance use history}} \\
\quad Benzodiazepine(s)
  & 211 & 182 & 0.863 & 0.924 & 176 & 0.834 & 0.906 & 120 & 0.569 & 0.717 \\
\quad Cocaine (including Crack)
  & 1609 & 1595 & 0.991 & 0.994 & 1564 & 0.972 & 0.980 & 1295 & 0.805 & 0.854 \\
\quad Injection drug use
  & 523 & 496 & 0.948 & 0.971 & 396 & 0.757 & 0.850 & 140 & 0.268 & 0.399 \\
\quad Ketamine
  & 27 & 27 & 1.000 & 1.000 & 26 & 0.963 & 0.981 & 20 & 0.741 & 0.850 \\
\quad Opioid(s)
  & 1638 & 1594 & 0.973 & 0.981 & 1586 & 0.968 & 0.977 & 1252 & 0.764 & 0.821 \\
\quad PCP
  & 53 & 50 & 0.943 & 0.971 & 47 & 0.887 & 0.939 & 36 & 0.679 & 0.808 \\
\quad Prescription medication(s)
  & 217 & 145 & 0.668 & 0.795 & 94 & 0.433 & 0.595 & 25 & 0.115 & 0.200 \\
\quad Methamphetamine / other stimulant
  & 141 & 135 & 0.957 & 0.978 & 130 & 0.922 & 0.958 & 100 & 0.709 & 0.826 \\
\quad Unspecified / other
  & 3715 & 3107 & 0.836 & 0.774 & 3333 & 0.897 & 0.854 & 1664 & 0.448 & 0.352 \\
\quad Treatment history
  & 834 & 783 & 0.939 & 0.963 & 792 & 0.950 & 0.970 & 544 & 0.652 & 0.761 \\
\quad History of prior overdose(s)
  & 502 & 443 & 0.882 & 0.932 & 426 & 0.849 & 0.911 & 329 & 0.655 & 0.776 \\
\midrule
\multicolumn{11}{@{}l}{\textit{Drugs / paraphernalia on scene}} \\
\quad Injection equipment
  & 559 & 556 & 0.995 & 0.997 & 551 & 0.986 & 0.992 & 491 & 0.878 & 0.929 \\
\quad Other paraphernalia$^{*}$
  & 2229 & 858 & 0.385 & 0.429 & 774 & 0.347 & 0.390 & 260 & 0.117 & 0.137 \\
\quad Packaging
  & 1120 & 1113 & 0.994 & 0.996 & 1117 & 0.997 & 0.998 & 1017 & 0.908 & 0.940 \\
\quad Pills / capsules
  & 422 & 363 & 0.860 & 0.919 & 395 & 0.936 & 0.964 & 137 & 0.325 & 0.471 \\
\quad Powder / residue
  & 1239 & 1225 & 0.989 & 0.993 & 1227 & 0.990 & 0.994 & 1064 & 0.859 & 0.904 \\
\quad Smoking equipment
  & 995 & 993 & 0.998 & 0.999 & 993 & 0.998 & 0.999 & 922 & 0.927 & 0.954 \\
\quad Straw or similar
  & 822 & 815 & 0.991 & 0.995 & 817 & 0.994 & 0.996 & 774 & 0.942 & 0.965 \\
\midrule
\textbf{Overall}\textsuperscript{\dag}
  & & & \textbf{0.976} & \textbf{0.910}
  & & \textbf{0.976} & \textbf{0.909}
  & & \textbf{0.934} & \textbf{0.718} \\
\bottomrule
\end{tabular}}
\end{table*}

\section{Comparison with Logistic regression}
We also see whether the use of language models is even necessary for the second stage $R_G \rightarrow O$ and whether simple models like logistic regression are sufficient to make the prediction. We report three logistic regression baselines across three experiments. Experiment~1 trains on human-annotated
evidence variables from the \spotlc\ split; because these annotations exist
only for \spotlc\ cases, it can be evaluated on the \spotlc\ test subset alone.
Experiment~2 trains on model-extracted evidence variables using the same
train/test case IDs as Experiment~1 to directly compare with human annotations, but can be evaluated across all
subsets. Experiment~3 trains on model-extracted variables pooled across all
splits. All models use balanced class weights. Table~\ref{tab:lr-macrof1}
reports macro-F1. We find that the performance remains subpar as compared to using language models and hence the context in the evidence and the report is required to actually get the final prediction. 

\section{Analysis where the evidence extracted is correct but the final prediction is not}
For Llama-3.1-8B, we also analyze the cases where the upstream evidence extraction is correct for all the features, but there is still a discrepancy in the prediction. We find that the majority of the cases are correctly labeled. 

\resizebox{0.4\textwidth}{!}{%
\begin{tabular}{ll cc}
\toprule
& & \multicolumn{2}{c}{\textbf{Predicted}} \\
\cmidrule(lr){3-4}
& & Yes & No \\
\midrule
\multirow{2}{*}{\rotatebox[origin=c]{0}{\textbf{Actual}}}
  & Yes & 317 \, (77.7\%) & 12 \, (2.9\%) \\
  & No  & 39 \, (9.6\%)   & 40 \, (9.8\%) \\
\bottomrule
\end{tabular}
}

\section{Expert feedback types}\label{sec:expert-feedback}
An analysis of types of expert feedback is present in Table \ref{tab:error_taxonomy}.
\begin{table*}[t]
\centering
\small
\caption{Types of expert feedback on feature-guided rationales $R_G$ extracted by the student model. Numbers in parentheses give the count of edited samples per type.} \label{tab:error_taxonomy}
\resizebox{0.9\linewidth}{!}{
\begin{tabularx}{\textwidth}{@{}p{2.4cm}p{4.2cm}X@{}}
\toprule
\textbf{Type} & \textbf{Error type} & \textbf{Description} \\
\midrule
\multirow{6}{*}{\parbox{2.4cm}{Over-extraction\\(111)}}
 & On-scene evidence as history & Scene items cited as a history of use \\
 & Prescription as misuse & Prescribed medications or medication lists cited as misuse \\
 & Generic mention as a specific drug & Vague mentions of drug use mapped to a specific substance \\
 & Neighboring scene category & Scene item assigned to an adjacent scene feature \\
 & Indirect clinical sign & Clinical signs or hospital visits cited as direct evidence \\
 & Negation or out-of-scope substance & Denials, abstinence, tobacco, or marijuana cited as evidence \\
\midrule
\multirow{3}{*}{\parbox{2.4cm}{Missed evidence\\(70)}}
 & Synonyms and street names & Substance named by a synonym or street name \\
 & Drug inside a listed history & Substance named within a list in the medical history \\
 & Signs of treatment & Treatment programs or support meetings \\
\midrule
Boundary (23) & Fragment merging & Fragmented spans merged into full sentences \\
\bottomrule
\end{tabularx}
}
\end{table*}

\section{Additional Model Results}
\label{app:additional_models}

\begin{table}[H]
\centering
\small
\caption{Macro-F1 of additional prediction models.} \label{tab:additional-models}
\begin{tabular}{lcc}
\toprule
Input & ClinicalModernBERT & DeepSeek \\
\midrule
$I \rightarrow O$      & 0.78 & 0.82 \\
$R_G \rightarrow O$    & 0.71 & 0.77 \\
$I, R_G \rightarrow O$ & 0.81 & 0.83 \\
\bottomrule
\end{tabular}
\end{table}

\begin{table}[H]
\centering
\small
\caption{Macro-F1 for logistic regression baselines across evidence variable source and training distribution. Exp.~1: \spot\ train samples trained with human evidence variables.
Exp.~2:  \spot\ train samples trained with model-extracted variables.
Exp.~3: Entire train split trained with model-extracted variables.
Dashes indicate evaluations unavailable by construction.} \label{tab:lr-macrof1}
\begin{threeparttable}
\begin{tabular}{l ccc}
\toprule
\multirow{2}{*}{\textbf{Test subset}}
  & \multicolumn{3}{c}{\textbf{Macro-F1}} \\
\cmidrule(lr){2-4}
 & Exp.~1 & Exp.~2 & Exp.~3 \\
\midrule
\spot\              & 0.552 & 0.561 & 0.523 \\
\nonspotrandom\           & ---   & 0.485 & 0.478 \\
All                 & ---   & 0.545 & 0.621 \\
\bottomrule
\end{tabular}
\end{threeparttable}
\end{table}

\end{document}